\documentclass[journal,twoside]{IEEEtran}

\usepackage{cite}

\usepackage{graphicx}
\usepackage{amsmath}

\usepackage{array}
\makeatletter		

\newcommand{\Rmnum}[1]{\expandafter\@slowromancap\romannumeral #1@}
\makeatother

\usepackage{algorithm}

\usepackage{multirow}
\usepackage{amssymb}
\usepackage{bm}		
\usepackage{xcolor}
\usepackage[colorlinks, citecolor=blue]{hyperref} 
\usepackage{threeparttable}
\usepackage{subfigure}
\usepackage{algorithmicx}
\usepackage{algpseudocode}
\usepackage{array}
\usepackage{textcomp}
\usepackage{stfloats}
\usepackage{url}
\usepackage{verbatim}
\usepackage{graphicx}
\usepackage{tabularx}

\newcolumntype{L}[1]{>{\raggedright\arraybackslash}p{#1}}
\newcolumntype{C}[1]{>{\centering\arraybackslash}p{#1}}
\newcolumntype{R}[1]{>{\raggedleft\arraybackslash}p{#1}}

\renewcommand{\algorithmicrequire}{\textbf{Input:}}
\renewcommand{\algorithmicensure}{\textbf{Output:}}

\usepackage{float}
\usepackage{color}
\usepackage{caption}
\usepackage{bbding}
\usepackage{arydshln}
\usepackage{booktabs}
\usepackage{colortbl}
\definecolor{Gray}{rgb}{0.5, 0.5, 0.5}
\usepackage{color}
\usepackage{amssymb}
\usepackage{amsmath}
\usepackage{bm}
\usepackage{makecell}
\usepackage[switch]{lineno}

\begin{document}
%
\title{MASA: Motion-aware Masked Autoencoder with Semantic Alignment for Sign Language Recognition}
%
%

\author{Weichao Zhao, Hezhen Hu, Wengang Zhou~\IEEEmembership{Senior Member,~IEEE}, Yunyao Mao, \\ Min Wang, Houqiang Li~\IEEEmembership{Fellow,~IEEE}
\thanks{This work is supported by National Natural Science Foundation of China under Contract U20A20183 \& 62021001, and the Youth Innovation Promotion Association CAS. It was also supported by the GPU cluster built by MCC Lab of Information Science and Technology Institution, USTC, and the Supercomputing Center of the USTC. }
\thanks{Weichao Zhao, Hezhen Hu, Wengang Zhou, Yunyao Mao and Houqiang Li are with the CAS Key Laboratory of Technology in Geospatial Information Processing and Application System, Department of Electronic Engineering and Information Science, University of Science and Technology of China, Hefei, 230027, China (e-mail: \{saruka, alexhu, myy2016\}@mail.ustc.edu.cn, \{zhwg, lihq\}@ustc.edu.cn.}
\thanks{Min Wang is with the Institute of Artificial Intelligence, Hefei Comprehensive National Science Center, Hefei 230030, China (e-mail: wangmin@iai.ustc.edu.cn).}
\thanks{Corresponding authors: Wengang Zhou and Houqiang Li.}}

%
%


\markboth{IEEE TRANSACTIONS ON CIRCUITS AND SYSTEMS FOR VIDEO TECHNOLOGY, June~2024}
{Motion-aware Masked Autoencoder with semantic alignment for Sign Language Recognition}

%



\maketitle

\begin{abstract}
    Sign language recognition~(SLR) has long been plagued by insufficient model representation capabilities.
Although current pre-training approaches have alleviated this dilemma to some extent and yielded promising performance by employing various pretext tasks on sign pose data, these methods still suffer from two primary limitations:
	\romannumeral 1) Explicit motion information is usually disregarded in previous pretext tasks, leading to partial information loss and limited representation capability.
	\romannumeral 2) Previous methods focus on the local context of a sign pose sequence, without incorporating the guidance of the global meaning of lexical signs.
	To this end, we propose a \textbf{M}otion-\textbf{A}ware masked autoencoder with \textbf{S}emantic \textbf{A}lignment~(MASA) that integrates rich motion cues and global semantic information in a self-supervised learning paradigm for SLR.
	Our framework contains two crucial components, \emph{i.e.,} a motion-aware masked autoencoder~(MA) and a momentum semantic alignment module~(SA).
	Specifically, in MA, we introduce an autoencoder architecture with a motion-aware masked strategy to reconstruct motion residuals of masked frames, thereby explicitly exploring dynamic motion cues among sign pose sequences.
	Moreover, in SA, we embed our framework with global semantic awareness by aligning the embeddings of different augmented samples from the input sequence in the shared latent space.
	In this way, our framework can simultaneously learn local motion cues and global semantic features for comprehensive sign language representation. 
	Furthermore, we conduct extensive experiments to validate the effectiveness of our method, achieving new state-of-the-art performance on four public benchmarks.
    The source code are publicly available at \href{https://github.com/sakura2233565548/MASA}{https://github.com/sakura/MASA}.
\end{abstract}


\ifCLASSOPTIONpeerreview
\begin{center} 
	\bfseries EDICS Category: 3-BBND 
\end{center}
\fi
%
\IEEEpeerreviewmaketitle

\section{Introduction}

Sign language serves as the primary communication tool within the deaf community.
Different from spoken language, sign language is characterized by its special grammar and lexicon, which causes difficulties in comprehension for hearing people. 
To facilitate communication between the deaf and hearing communities, sign language recognition~(SLR) has received significant attention for its potential social impact.
As a fundamental task, isolated SLR aims to recognize the meaning of sign language video at the gloss-level under complex situations, \textit{i.e.,} quick motion, unconstrained background and individual diversities. In this work, we focus on isolated SLR.

Most existing isolated SLR works~\cite{li2020word,li2020transferring,koller2018deep,hu2023continuous,joze2018ms,tunga2021pose,selvaraj-etal-2022-openhands,9393618, 10105511} are directly optimized on the target benchmark in a supervised learning paradigm, which is prone to over-fitting due to insufficient sign data resources. 
To address this issue, some methods~\cite{hu2021signbert,zhao2023best,albanie2020bsl,jiang2021skeletor,hu2023signbert+} attempt to design different pre-training strategies to boost performance on isolated SLR, inspired by the related progress in NLP and CV fields. 
Among them, Albanie \textit{et al.}~\cite{albanie2020bsl} utilize the pose keypoints as supervision signals to force the model to pay more attention to the foreground signer from RGB images. 
Considering that pose is a more compact representation than RGB images, SignBERT~\cite{hu2021signbert} introduces a self-supervised pre-trainable framework with available sign pose data.
It captures the contextual information by masking and reconstructing hand joints among a sign pose sequence and achieves promising performance.
Additionally, BEST~\cite{zhao2023best} tokenizes continuous joints into a discrete space as the pseudo labels and performs similar pre-training as BERT~\cite{devlin2018bert} via reconstructing the masked tokens from the corrupted input sequence.  

Despite the decent results made by the above methods, there still exist several unresolved issues.
Firstly, motion dynamics are widely regarded as a crucial role in the field of sign language understanding~\cite{Mercanoglu_Sincan_2022,saunders2021mixed,liang2021dynamic}.
However, existing pre-training methods set the objective as recovering the static information (e.g. joint spatial location), which constrains its modeling capability on motion cues among frames and leads to sub-optimal performance.   
Secondly, these methods do not explicitly incorporate the global meaning present in sign pose data.
Since the meaning of a lexical sign is generally expressed by a series of poses, the instance representation of a sign pose sequence also holds effective information for sign language comprehension.
Unfortunately, previous methods~\cite{hu2021signbert,zhao2023best,jiang2021skeletor} mainly focus on learning local contextual information for fulfilling the reconstruction task, while lacking the explicit exploration of such crucial representations.

Inspired by the above observations, we propose a novel motion-aware masked autoencoder with semantic alignment via a self-supervised learning paradigm for SLR.
Our framework aims to leverage effective motion information and global contextual meaning to learn more effective representation during pre-training.
On one hand, in order to explicitly mine motion cues among the sign pose sequence, we introduce a motion-aware masked autoencoder~(MA).
Concretely, we present a motion-aware masked strategy, which purposely selects the regions with the conspicuous motion for masking.
Given the masked sign pose sequence, we utilize the remaining information to reconstruct the motion residuals of masked frames rather than the static joints.
By combining these key designs, our proposed method can focus on more informative regions for capturing dynamic pose variations and improving learned representations.

On the other hand, we design a momentum semantic alignment module~(SA) to endow our model with the capability of learning consistent instance representation. 
Specifically, we generate two augmented samples from the input sequence, treating them as positive samples and aligning the embeddings of positive samples in the shared latent space with a contrastive objective.  
Additionally, we incorporate a momentum mechanism to update the parameters of most models in SA, reducing extra computation costs.
In this way, our proposed method enables the explicit exploitation of motion cues and global semantic meaning to jointly learn powerful representations. 

The key contributions of our work are summarized as follows:
\begin{itemize}
	\item We propose a novel motion-aware masked autoencoder with semantic alignment for sign language recognition. It aims to leverage explicit motion information and global semantic meaning in sign pose data to fertilize sign language representation learning. 
	\item We introduce two main components in our framework, \textit{i.e.,} motion-aware masked autoencoder~(MA) and momentum semantic alignment module~(SA). In MA, we delicately design masked strategy and an optimized objective to effectively capture dynamic motion cues among sign pose sequence. In SA, we introduce a module to guide our framework learning the globally consistent meaning of lexical signs by aligning features in the latent space. In this way, our model is endowed with capturing both local pose variation and global sign meaning for more holistic representations. 
	\item Extensive experiments on isolated SLR validate the effectiveness of our proposed method. Compared with previous methods, our method achieves new state-of-the-art performance on four public benchmarks, \textit{i.e.,} WLASL, MSASL, NMFs\_CSL and SLR500.
\end{itemize}

\section{Related Work}
In this section, we will briefly review several related topics for this paper, including sign language recognition and self-supervised representation learning.

\subsection{Sign Language Recognition} 
Sign language recognition as a fundamental task has received increasing attention in recent years~\cite{koller2018deep,rastgoo2021sign,koller2020quantitative}.
With the popularity of video capture devices, the majority of methods utilize RGB videos as their research target.
Sarhan~\textit{et al.,}~\cite{sarhan2023unraveling} present a comprehensive survey on the development of ISLR research, covering various aspects including modalities, parameters, fusion categories and transfer learning.
Generally, these research works are grouped into two categories based on the input modality, \textit{i.e.,} RGB-based and pose-based methods.

\noindent \textbf{RGB-based Methods.} 
Early studies in SLR predominantly focused on the design of hand-crafted features to represent hand shape variation and body motion~\cite{farhadi2007transfer,fillbrandt2003extraction,starner1995visual,6704715}.
However, with the advance of deep learning techniques, there has been a notable shift in the field, with a growing number of works adopting convolutional neural networks~(CNNs) as the fundamental architecture for feature extraction~\cite{hu2021global,hu2021hand,koller2018deep,sincan2020autsl,selvaraj-etal-2022-openhands,laines2023isolated, 9528010, 8707080}. Since the input signal for SLR is an RGB video sequence, some works~\cite{hu2021global,li2020transferring,9106402,1318645,10185608} employ 
2D CNN networks to extract frame-wise features and then aggregate temporal information to predict sign language. 
To better model spatial-temporal representations in sign language videos, some works~\cite{huang2018attention,joze2018ms,albanie2020bsl,li2020transferring,li2020word} attempt to adopt 3D CNNs,~\textit{i.e.,} I3D~\cite{carreira2017quo} and S3D~\cite{xie2018rethinking} to extract video-level features due to their capacity of spatial-temporal dependency.
NLA-SLR~\cite{zuo2023natural} leverages RGB videos and the human keypoints heatmaps with different temporal receptive fields to build a four-stream framework based on S3D~\cite{xie2018rethinking}, which improves the recognition accuracy through complex model design. Bilge~\textit{et al.,}~\cite{bilge2022towards} propose a zero-shot learning framework on ISLR task, which leverages the descriptive text
and attribute embeddings to transfer knowledge to the instances of unseen sign classes. 
	
\noindent \textbf{Pose-based Methods.}
Compared to RGB images, human pose includes the meaningful expressions conveyed in sign language videos through human skeleton movements, while reducing interference factors like complex backgrounds and external appearances of signers.
Due to naturally physical connections among skeleton joints, graph convolutional networks~(GCNs) are widely applied to model spatial-temporal relations as the backbone~\cite{li2018co,du2015hierarchical,yan2018spatial, li2020word,jiang2021skeletor,ng2021body2hands,Lee_2023_ICCV}.
Kindiroglu~\textit{et al.,}~\cite{kindiroglu2024transfer} explore the feasibility of transferring knowledge among different benchmarks with GCN-based approaches. TSSI~\cite{laines2023isolated} converts a skeleton sequence into an RGB image and utilizes an RGB-based backbone to model sequential skeleton features, achieving promising improvement.
Hu \textit{et al.}~\cite{hu2021signbert, hu2023signbert+} propose a self-supervised learning framework SignBERT by masking and reconstructing hand pose sequence to learn local contextual information.
To better leverage the success of BERT into the sign language domain, Zhao \textit{et al.}~\cite{zhao2023best} tokenize sign pose into a triplet unit in a frame-wise manner and also introduce a pre-trainable framework namely BEST to improve the performance of sign language recognition. However, these pre-training methods overlook the importance of dynamic motion information, while our proposed method explicitly explores motion cues among sign pose sequences to learn more effective representation.

\subsection{Self-supervised Representation Learning}
Self-supervised representation learning aims to learn effective representation from massive unlabeled data, which shows better generalizability in downstream tasks.
To supervise the pre-training process, researchers design variant pretext tasks, \textit{i.e.,} jigsaw puzzles~\cite{noroozi2016unsupervised,noroozi2018boosting}, colorization~\cite{zhang2016colorful}, predicting rotation~\cite{gidaris2018unsupervised}, masked patch reconstruction~\cite{He_2022_CVPR,tong2022videomae,huang2022contrastive}, contrast similarity between samples~\cite{chen2020simple,he2020momentum,oord2018representation,qian2021spatiotemporal}, \textit{etc.}
Recently, masked patches reconstruction and contrastive learning have become mainstream methods in video representation learning.
The former focuses more on the local relations in the input video, while the latter inclines to model relations among different videos. 
Typically, VideoMAE~\cite{tong2022videomae} proposes customized video tube masking to enforce high-level structure learning and obtains impressive performance.
In contrast, VideoMoCo~\cite{pan2021videomoco} learns instance discriminative representation by encoding similar features from two perspectives in an input video and pulling the distance of similar features.

In order to strengthen video representation learning, a few works~\cite{tao2022siamese,zhou2021ibot,huang2022contrastive,wang2023molo, baradel2021leveraging, zhu2022motionbert,mao2023masked} attempt to leverage both prevailing techniques. 
In fields related to human behavior, PoseBERT~\cite{baradel2021leveraging} utilizes a masked modeling technique on the pose parameters of the human model SMPL, which aims to improve the performance of human mesh recovery.
Mao~\textit{et al.,}~\cite{mao2023masked} propose a masked motion predictor to better capture contextual information among human pose sequences.
Conversely, MotionBERT~\cite{zhu2022motionbert} leverages a substantial volume of human motion data captured through motion capture devices to improve downstream tasks.
In the sign language domain, current self-supervised methods~\cite{hu2021signbert,zhao2023best,jiang2021skeletor} are based on masked modeling strategy and overlook the discriminative representation of a lexical sign, which is also essential to sign language understanding. 
To address this problem, we first explore to perceive comprehensive global information in local frame features via self-supervised learning.

\begin{figure*}[t]
	\centering
	\includegraphics[width=1.0\textwidth]{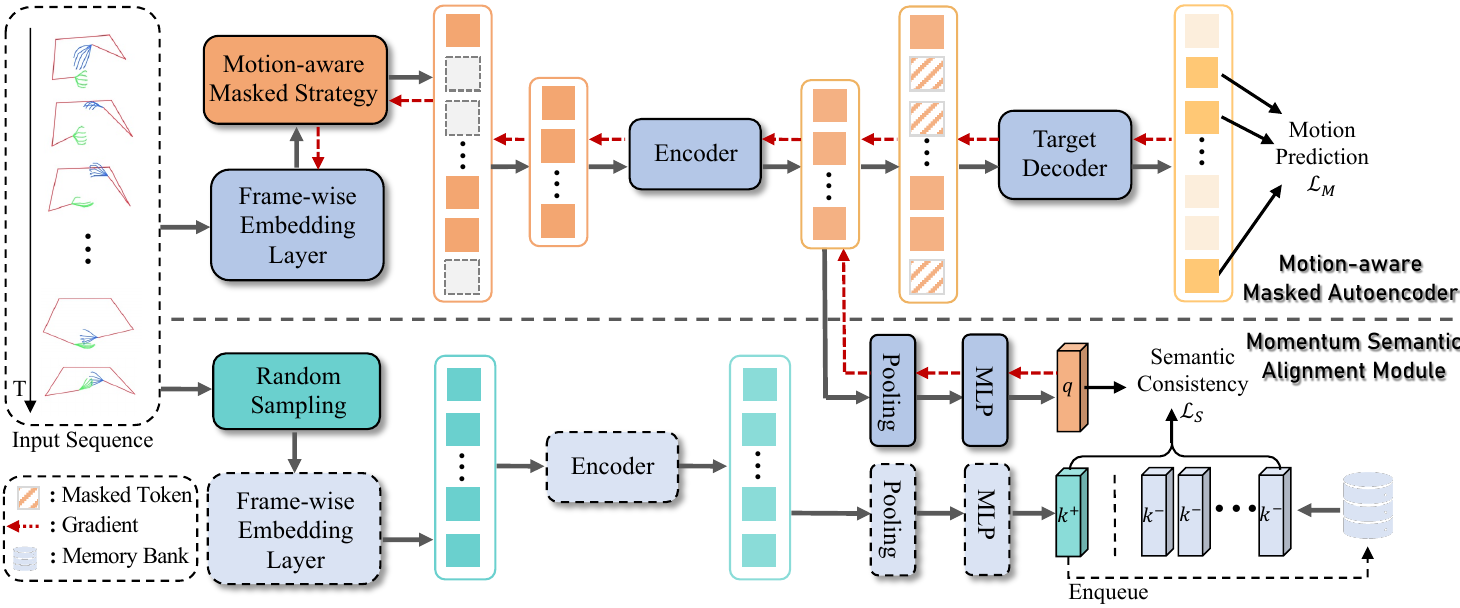} 
	\caption{The illustration of our proposed framework during pre-training. It mainly contains two crucial components, \textit{i.e.,} a motion-aware masked autoencoder~(MA) and a momentum semantic alignment module~(SA). In MA, given a sign pose sequence, the frame-wise embedding layer encodes each frame pose into a latent feature. The encoder operates on unmasked feature sequences. Then, the target decoder predicts the motion residuals from pose tokens~(along with mask tokens). In SA, it employs a siamese framework extracting pose features from randomly sampled pose sequence similar to MA. After that, we project global features from positive views into a shared embedding space and align them with the semantic consistency loss. The modules with the black dashed box are momentum updated and the red arrows represent the gradient back propagation.}
	\label{fig1}
\end{figure*}

\section{Method}
The overall of our proposed method is illustrated in Fig.~\ref{fig1}, which consists of a motion-aware masked autoencoder~(MA) and a momentum semantic alignment module~(SA). 
In MA, we present a novel motion-aware masked strategy to prioritize more informative regions for masking, avoiding randomly masking invalid frames, \textit{i.e,} consecutive similar poses and low-confidence regions.
In SA, we extract two sequence-level representations from augmented samples of the input sequence. 
Then, we impose constraints on the consistency of both features in a latent embedding space to fertilize our model with discriminative capability. 
In the following, we first elaborate these components separately. Then, we present our training objective. Finally, the downstream fine-tuning of our approach is discussed. Conveniently, we provide a terminology table in Tab.~\ref{terminology} for better understanding of our method.

\begin{table}[t]
    \footnotesize
    \tabcolsep=10.0pt
    \caption{The terminology table of the main symbols utilized in our proposed method.}
    \label{terminology}
    \begin{center}
        \resizebox{0.9\linewidth}{!}{
            \begin{tabular}{cc}
                \toprule
                Symbol & Description \\
                \midrule
                $V_{in}$ & Input pose sequence \\
                $R_{in}$ &Input pose confidence \\
                $M$ & Motion residual of input pose $V_{in}$ \\
                $C$ & High confidence of input confidence $R_{in}$ \\
                $\mathcal{M}$ & Selected masked index set \\
                $F_e$ & Pose embedding sequence \\
                $\tilde{F}_e$ & Unmasked part of pose embedding sequence $F_e$ \\
                $F_{mask}$ & Embedding sequence of learnable masked tokens \\
                $V_{aug}$ & Augmented sample of input pose sequence $V_{in}$ \\
                $\alpha, \alpha_r$ & Proportion for randomly masking \\
                \bottomrule
                
            \end{tabular}
        }
    \end{center}
\end{table}

\subsection{Motion-aware Masked Autoencoder}

\noindent \textbf{Motion-aware Masked Strategy.}  Given the sign pose sequence $\mathit{V}_{in} = \{x_i \in \mathbb{R}^{N \times 2}\}_{i=1}^T$, we first compute the motion residuals among these frames denoted as $\mathit{M} = \{m_i \ | \ x_{i+k} - x_i \ , \ if \ 0 \leq i \leq T-k \ , else \ \mathop{\bold{0}}\limits ^{\rightarrow} \}_{i=1}^T$, in which $k$ represents the interval length between two frames.
Since the hand motion is inherently coupled with the upper body that leads to difficultly capturing fine-grained hand movements, we crop hand and upper body parts separately to emphasize these dominant regions and then compute the movements of each part. 
Based on the above operation, $m_i$ contains the decoupled pose variations of both hands and the upper body at the $i$-th timestep.

As the pose sequence is estimated utilizing an off-the-shelf estimator, the resulting confidence values for each pose provide essential information regarding the reliability of the estimation process. 
Thus, we organize the original confidence values of $\mathit{V}_{in}$ as $\mathit{R}_{in}= \{r_i \in \mathbb{R}^{N}\}_{i=1}^T$.
Moreover, in order to eliminate the interference of low-confidence joints of each pose, we set a threshold $\epsilon_c$ to truncate low confidence values, denoted as  $\mathit{C}= \{c_i:= r_i \cdot \mathbb{I}(r_i \geq \epsilon_c) \}_{i=1}^T$, where $\mathbb{I}(\cdot)$ is an indicator function.
Then, given the motion sequence $M$ and the pose confidence $C$, we could obtain the high-confidence motion information and choose the frames with conspicuous motion into the candidate set, in which a fraction of frames is randomly selected for masking, as shown in Alg.~\ref{alg1}.
Finally, we can obtain the selected masked index set $\mathcal{M}$.

\noindent \textbf{Frame-wise Pose Embedding Layer.} We utilize this embedding layer to embed each frame pose into the latent space. Motivated by SignBERT~\cite{hu2021signbert}, we design a graph convolutional network~(GCN) comprising two sub-networks modified from~\cite{cai2019exploiting}, which aims to extract the latent embeddings of both hands and upper body respectively. Specifically, given the sign pose sequence $V_{in}$ of T frames, the latent embedding of the $i$-th frame is extracted as follows,
\begin{equation}
	\begin{split}
		f_i &= Embed\_Layer(x_i) \\ &= Concat(\mathrm{GCN\_H}(x_i^h), \ \mathrm{GCN\_B}(x_i^{b})), \quad x_i \in \mathit{V}_{in},
	\end{split}
\end{equation} 
where $\mathrm{GCN\_H(\cdot)}$ and $\mathrm{GCN\_B(\cdot)}$ denote two sub-networks, while $x_i^h$ and $x_i^b$ represent the hand and body parts of $x_i$, respectively.
Thus, the embedding sequence is denoted as $\mathit{F}_{e} = \{ f_i \in \mathbb{R}^{D_e} \}_{i=1}^T$.

\begin{algorithm}[t]
	\renewcommand{\algorithmicrequire}{\textbf{Input:}}
	\renewcommand{\algorithmicensure}{\textbf{Output:}}
	\caption{Motion-aware Masked Strategy}
	\label{alg1}
	\begin{algorithmic}[1]
		\Require $M = \{m_i\}_{i=1}^T$; $\ C = \{c_i\}_{i=1}^T$; $\mathcal{S} = \{\}$ 
		\For {$i \leftarrow 1$ \textbf{to} $T-k$} \Comment{$k$ is the interval length.}
		\State $\widetilde{c} = c_i \cdot c_{i+k}$
		\State $\widetilde{m}_{i} = |m_i| \cdot \widetilde{c}$ \Comment{Truncate low-confidence motion movements.}
		\State $p_i = $ sum($\widetilde{m}_{i} \geq \epsilon_m$) / sum($\widetilde{m}_{i} \geq 0$) \Comment{Compute the ratio of keypoints with valid motion.}
		\If{$p_i \geq \delta$}
		\State $\mathcal{S}$.append($i$) \Comment{Add the corresponding index into the candidate set.}
		\EndIf
		\EndFor
		\State $\mathcal{M} = Random\_Select(\mathcal{S}, \alpha)$ \Comment{Randomly select a fixed proportion $\alpha$ of elements in $\mathcal{S}$.}
		\Ensure $\mathcal{M}$
	\end{algorithmic}
\end{algorithm}

\noindent \textbf{Encoder.} Given the embedding sequence $\mathit{F}_e$ and the masked index set $\mathcal{M}$, we split $\mathit{F}_e$ into two parts, \textit{i.e.,} the unmasked sequence $\widetilde{F}_e = \{f_i | i \not\in \mathcal{M} \}$ and a series of learnable masked token $\mathit{F}_{mask} = \{ \mathbf{e}_{i,mask} \in \mathbb{R}^{D_e} | i \in \mathcal{M} \}$. The encoder adopts the vanilla Transformer~\cite{vaswani2017attention} architecture and is applied to $\widetilde{F}_e$. We feed them into a sequence of transformer blocks and obtain the output sequence $\mathit{F}_{enc} = Encoder(\widetilde{F}_e)$, which is utilized to reconstruct masked information.

\noindent \textbf{Target Decoder.} The target decoder aims to recover the masked pose information from the visible features. 
Specifically, the decoder receives both the encoder feature sequence $\mathit{F}_{enc}$ and the masked tokens $\mathit{F}_{mask}$ as input.
We first concatenate  $\mathit{F}_{enc}$ and $\mathit{F}_{mask}$ together in the original order.
Then, another vanilla Transformer~\cite{vaswani2017attention} is utilized to mine effective contextual information from the encoder features as a decoder and complements missing information of masked tokens, which is computed as follows, 
\begin{equation}
	\mathit{F}_{dec} = Decoder(\mathit{F}_{enc}, \mathit{F}_{mask}),
\end{equation}
where $\mathit{F}_{dec} = \{ f_{i, dec} \in \mathbb{R}^{D_e} \}_{i=1}^T$ denotes the output of stacked transformer blocks.
Finally, a simple MLP predicts the motion residuals of masked frames using the corresponding recovered features, which is computed as follows,
\begin{equation}
	\quad y_{i} = \mathbb{I}(i \in \mathcal{M}) \cdot \mathcal{P}(f_{i, dec}), \quad \mathcal{Y} = \{ y_{i} | i \in \mathcal{M}\},
\end{equation}
where $\mathbb{I}(i \in \mathcal{M})$ is an indicator to only select the predictions corresponding to masked tokens, and $\mathcal{Y}$ is the output prediction for the masked frames.

\noindent \textbf{Motion Prediction.} Given the output of the target decoder $\mathcal{Y}$, we supervise it with the motion residuals $\mathit{M}$. The motion prediction loss $\mathcal{L}_m$ is computed as follows,
\begin{equation}
	\mathcal{L}_m = \frac{1}{L} \sum_{i \in \mathcal{M}} || (y_{i} - m_{i}) \cdot c_i \cdot c_{i+k} ||^2, \quad m_i \in \mathit{M}, c_i \in \mathit{C}, 
\end{equation}
where $L$ is the size of the masked index set $\mathcal{M}$. Since the ground-truth motion residual is inferred from the estimated sign poses, it is pertinent to incorporate the confidence values of the corresponding sign poses into the objective function.

\subsection{Momentum Semantic Alignment Module}
\noindent In MA, our framework aims to reconstruct the partially masked information via the mining remaining sign pose features, which prefers to focus on local motion contexts, without considering the global representation of sign pose sequence.
Thus, we design this module~(SA) to endow our framework with global instance discriminability by aligning the semantic features in the shared common space.
In SA, we adopt the same architectures with MA,~\textit{i.e.,} the frame-wise embedding layer and the encoder, while incorporating another augmented sequence as input. 
As mentioned in MoCo~\cite{he2020momentum}, data augmentation is essential to learn effective global representation. If we preserve complete temporal information in the positive sample, the model could be easily converged to a sub-optimal solution due to the simple pretext task.
To this end, we utilize a temporal augmentation to generate proper positive samples, which not only considers the design of MA, but also avoids the large disparity between different augmented samples to hamper the learning of global comprehensive information.

Considering that the masked strategy in MA preserves less information in the corresponding visible sequence, we utilize a random temporal sampling strategy to make a trade-off between the availability of information and the complexity of  data augmentation, which is computed as follows,
\begin{equation}
	\mathit{V}_{aug} = Random\_Sample(\mathit{V}_{in}, \alpha_r),
\end{equation}
where $\alpha_r$ denotes the proportion for randomly masking, and $\mathit{V}_{aug}$ includes $(1 -\alpha_r) \cdot T$ frames as the positive sample.
Then, we encode it through a similar pipeline to obtain the corresponding latent feature~$\mathit{F}_{enc}^+$. 
In this way, the latent features of query and its positive key could be obtained from the encoder in MA and SA, which are denoted as $\mathit{F}_{enc}$ and $\mathit{F}_{enc}^+$, respectively.
Next, we extract the global representation of each sample using an average pooling layer.
To better align the representations of positive samples, we further utilize two projection heads to transform them into a shared embedding space, which is computed as follows,

\begin{align}
    q & = Proj\_q(AvgPool(\mathit{F}_{enc})), \\ 
    \quad k^+ & = Proj\_k(AvgPool(\mathit{F}_{enc}^+)),
\end{align}

\noindent where $q$ and $k^+$ denote the embedding features of query and key samples, respectively. Similarly, the embedding features from other sign pose sequences are considered as negative keys, denoted as $k^-$.

Finally, a semantic consistency loss is adopted to force the model to learn invariant instance representation by pulling close the embeddings of the query and positive key generated from different augmented samples and pushing away negative keys from the other sequences. 
Similar to MoCo~\cite{he2020momentum}, a memory bank $\mathcal{B}$ is utilized to store the key embeddings in each training step in a first-in-first-out~(FIFO) manner and provides the negative keys for the next steps.
Based on the above settings, the semantic consistency loss is derived from the InfoNCE loss~\cite{oord2018representation}, which is computed as follows,
\begin{equation}
	\mathcal{L}_s = -\mathrm{log}\frac{\mathrm{exp}(q^\top k^+ / \tau)}{\mathrm{exp}(q^\top k^+ / \tau) + \sum\nolimits_{i=1}^{K} \mathrm{exp}(q^\top k_{i}^- / \tau)}, \quad k_{i}^- \in \mathcal{B},
\end{equation}
where $\tau$ is the temperature coefficient, and $K$ is the size of the queue-based memory bank $\mathcal{B}$.
Moreover, we update the parameters of the architectures with the black dashed box in SA by exponential moving average~(EMA), as shown in Fig.~\ref{fig1}.
For instance, assuming that the parameters of the same modules of two branches denote as $\theta_q$ and $\theta_k$,  $\theta_k$ is updated by $ \theta_k \leftarrow \mu \theta_k + (1-\mu) \theta_q$.  The hyper-parameter $\mu$ is fixed as 0.996 in our experiment. Momentum update is widely utilized due to its stability during training by fostering smooth feature changes.

\subsection{Training Objective}
During the pre-training stage, the overall objective is the combination of  local motion prediction objective  and global semantic consistency objective, which is computed as follows,
\begin{equation}
	\mathcal{L} = \mathcal{L}_{m} + \lambda_s \mathcal{L}_s , 
\end{equation}
where $\lambda_s$ is a loss weight controlling the relative influence of both losses. In practice, we utilize a schedule to gradually introduce the global semantic consistency objective and guide the model to learn suitable local feature at the beginning of training.

\subsection{Downstream Fine-Tuning}
After pre-training, we only utilize the partial pre-training framework, including a frame-wise embedding layer, an encoder and a pooling layer. Then, a simple classifier is attached after them for SLR task. During fine-tuning, we sample the fixed number of frames as input.
We adopt the common 
Following~\cite{hu2021signbert}, we utilize random and center temporal sampling during training and testing, respectively. In our experiment, we report our results as \textit{Ours}. Moreover, we utilize a late fusion strategy mentioned in~\cite{hu2021signbert} to enhance the RGB-based method with the prediction results of our method, namely \textit{Ours}~(+R).

\section{Experiment}

\subsection{Implementation Details}
\noindent \textbf{Data Preparation.} We utilize sign pose data to represent the hand and body information, which contains 2D pose keypoints extracted from the off-the-shelf estimator MMPose~\cite{mmpose2020}. In each frame, the pose information includes $N=49$ joints, containing 7 body joints and 42 hand joints with their corresponding confidences.

\noindent \textbf{Parameter Setup.} During pre-training stage, the dimension $D_e$ of frame-wise pose feature is 1536. In MA, the thresholds $\epsilon_c, \epsilon_m, \delta$ are set to 0.4, 5.0 and 0.5, respectively, and the masked ratio $\alpha$ is set to 90\%. In SA, we randomly mask $\alpha_r$ 50\% of the input sequence. The size $K$ of the memory bank is set to 6,144, and the temporal coefficient $\tau$ is set to 0.07. The loss weight $\lambda_s$ is set to 0.05 with a linear increment from zero in the first 100 epochs. We adopt AdamW~\cite{loshchilov2017decoupled} optimizer as default, and the momentum is set to $\beta_1 = 0.9$, $\beta_2 = 0.95$ with the weight decay set as 1e-4. The learning rate is set to 1e-4 with a warmup of 20 epochs and linear learning rate decay to train the overall framework for total 400 epochs. During the fine-tuning stage, we utilize the SGD~\cite{robbins1951stochastic} optimizer with 0.9 momentum. The learning rate is initialized to 0.01, then reduced by a factor of 0.1 every 20 epochs. We totally train the framework for 60 epochs with a batch size of 64. Following~\cite{hu2021signbert}, we sample 32 frames as input. All experiments are implemented by PyTorch on NVIDIA RTX 3090.

\begin{table*}[t]
	\footnotesize
	\tabcolsep=6.0pt
	\caption{Comparison with state-of-the-art methods on WLASL dataset. ``$\dagger$" indicates the model with self-supervised learning, and ``$\ast$" indicates the method utilized for late fusion. ``$\ddagger$" indicate the method that adopts 32 frames as input.}
	\begin{center}
		\resizebox{\textwidth}{!}{
			\begin{tabular}{lcccccccccccc}
				\toprule
				\multirow{3}{*}[-0.1in]{\makecell[c]{Methods}} & \multicolumn{4}{c}{WLASL100}
				& \multicolumn{4}{c}{WLASL300}
				& \multicolumn{4}{c}{WLASL2000}\\ 
				& \multicolumn{2}{c}{P-I} & \multicolumn{2}{c}{P-C} 
				& \multicolumn{2}{c}{P-I} & \multicolumn{2}{c}{P-C}
				& \multicolumn{2}{c}{P-I} & \multicolumn{2}{c}{P-C} \\ 
				\cmidrule(lr){2-3} \cmidrule(lr){4-5} \cmidrule(lr){6-7} \cmidrule(lr){8-9} \cmidrule(lr){10-11} \cmidrule(lr){12-13}
				& T-1 & T-5 & T-1 & \multicolumn{1}{c}{T-5}  
				& T-1 & T-5 & T-1 & \multicolumn{1}{c}{T-5} 
				& T-1 & T-5 & T-1 & \multicolumn{1}{c}{T-5}    \\ \toprule
				\textbf{Pose-based} & & & & & & & & & & & & \\
				ST-GCN~\cite{yan2018spatial}& 50.78 & 79.07 & 51.62 & 79.47   
				& 44.46 & 73.05 & 45.29 & 73.16
				& 34.40 & 66.57 & 32.53 & 65.45 \\ 
				Pose-TGCN~\cite{li2020word} & 55.43 & 78.68 & - & -   
				& 38.32 & 67.51 & - & -
				& 23.65 & 51.75 & - & -\\ 
				PSLR~\cite{tunga2021pose}& 60.15 & 83.98 & - & -   
				& 42.18 & 71.71 & - & -
				& - & - & - & - \\
				SignBERT~\cite{hu2021signbert}$^\dagger$ & 76.36 & 91.09 & 77.68 & 91.67  
				& 62.72 & 85.18 & 63.43 & 85.71 
				& 39.40 & 73.35 & 36.74 & 72.38  \\ 
				BEST~\cite{zhao2023best}$^\dagger$ & 77.91& 91.47 & 77.83 & 92.50 & 67.66 & 89.22 & 68.31 & 89.57 & 46.25 & 79.33 & 43.52 & 77.65 \\
                    NLA-SLR-K32~\cite{zuo2023natural}$^\ddagger$ & - & - &- & - & - & - & - & - & 46.66 & 79.95 & 43.81 & 78.49\\
				P3D~\cite{Lee_2023_ICCV} & 76.71 & 91.97 &  78.27 & 92.97 & 67.18 & 89/01 & 67.62 & 89.24 & 44.47 & 79.69 & 42.18 & 78.52 \\
				\textit{Ours}& \textbf{83.72} & \textbf{93.80} & \textbf{84.47} & \textbf{94.30} & \textbf{73.65} & \textbf{91.77} & \textbf{74.33} & \textbf{92.13} & \textbf{49.06} & \textbf{82.90}& \textbf{46.91} & \textbf{81.80} \\
				\midrule
				\textbf{RGB-based} & & & & & & & & & & & & \\
				I3D~\cite{carreira2017quo}$^\ast$ & 65.89 & 84.11 & 67.01 & 84.58  
				& 56.14 & 79.94 & 56.24 & 78.38
				& 32.48 & 57.31 & - & -\\ 
				TCK~\cite{li2020transferring} & 77.52 & 91.08 & 77.55 & 91.42   
				& 68.56 & 89.52 & 68.75 & 89.41
				& - & - & - & - \\ 
				BSL~\cite{albanie2020bsl} & - & - & - & -  
				& - & - & - & -
				& 46.82 & 79.36 & 44.72 & 78.47\\
				HMA~\cite{hu2021hand} & - & - & - & -
				& - & - & - & - 
				& 37.91 & 71.26 & 35.90 & 70.00 \\
                     NLA-SLR-VK32~\cite{zuo2023natural}$^\ddagger$ & - & - &- & - & - & - & - & - & 52.95 & 85.75 & 50.26 & 84.50 \\ 
				SignBERT~(+R)~\cite{hu2021signbert}   & 82.56 & 94.96 & 83.30 & 95.00
				& 74.40 & 91.32 & 75.27 & 91.72
				& 54.69 & 87.49 & 52.08 & 86.93 \\
				BEST~(+R)~\cite{zhao2023best} & 81.01 & 94.19 & 81.63 & 94.67 & 75.60 & 92.81 & 76.12 & 93.07 & 54.59 & 88.08 & 52.12 & 87.28 \\ 
				\textit{Ours}~(+R) & \textbf{84.88} & \textbf{95.74} & \textbf{85.63} & \textbf{95.92}  & \textbf{78.44} & \textbf{92.96} & \textbf{79.16} & \textbf{93.29}  & \textbf{55.77} & \textbf{88.85} & \textbf{53.13} & \textbf{88.14} \\
				\bottomrule 
			\end{tabular}
		}
	\end{center}
	\label{wlasl}
\end{table*}

\begin{table*}[t!]
	\footnotesize
	\tabcolsep=6.0pt
	\caption{Comparison with state-of-the-art methods on MSASL dataset. ``$\dagger$" indicates the model with self-supervised learning, and ``$\ast$" indicates the method utilized for late fusion. ``$\ddagger$" indicate the method that adopts 32 frames as input.}
	\begin{center}
		\resizebox{\textwidth}{!}{
			\begin{tabular}{lcccccccccccc}
				\toprule
				\multirow{3}{*}[-0.1in]{\makecell[c]{Methods}} & \multicolumn{4}{c}{MSASL100}
				& \multicolumn{4}{c}{MSASL200}
				& \multicolumn{4}{c}{MSASL1000}\\ 
				& \multicolumn{2}{c}{P-I} & \multicolumn{2}{c}{P-C} 
				& \multicolumn{2}{c}{P-I} & \multicolumn{2}{c}{P-C}
				& \multicolumn{2}{c}{P-I} & \multicolumn{2}{c}{P-C} \\ 
				\cmidrule(lr){2-3} \cmidrule(lr){4-5} \cmidrule(lr){6-7} \cmidrule(lr){8-9} \cmidrule(lr){10-11} \cmidrule(lr){12-13}
				& T-1 & T-5 & T-1 & \multicolumn{1}{c}{T-5}  
				& T-1 & T-5 & T-1 & \multicolumn{1}{c}{T-5} 
				& T-1 & T-5 & T-1 & \multicolumn{1}{c}{T-5}    \\ \toprule
				\textbf{Pose-based} & & & & & & & & & & & & \\
				ST-GCN~\cite{yan2018spatial} & 59.84 & 82.03 & 60.79 & 82.96  
				& 52.91 & 76.67 & 54.20 & 77.62
				& 36.03 & 59.92 & 32.32 & 57.15 \\ 
                    SL-TSSI~\cite{laines2023isolated} & 81.47 & - & - & - & - & - & - & - & - & - & - & - \\
                SignBERT~\cite{hu2021signbert}$^\dagger$ & 76.09 & 92.87 & 76.65 & 93.06 & 70.64 & 89.55 & 70.92 & 90.00
				& 49.54 & 74.11 & 46.39 & 72.65 \\
				BEST~\cite{zhao2023best}$^\dagger$ & 80.98 & 95.11 & 81.24 & 95.44 & 76.60 & 91.54 &76.75 &91.95 &58.82 &81.18& 54.87 & 80.05\\
                NLA-SLR-K32~\cite{zuo2023natural}$^\ddagger$ & - & - &- & - & - & - & - & - & 58.26 & 80.85 & 54.57 & 79.28\\
				\textit{Ours} & \textbf{83.22} & \textbf{95.24} & \textbf{83.19} & \textbf{95.46} & \textbf{79.25} &\textbf{92.86} &\textbf{79.70} &\textbf{93.33} & \textbf{63.47} &\textbf{83.89} & \textbf{60.79} & \textbf{83.29}\\
				\midrule
				\textbf{RGB-based} & & & & & & & & & & & &   \\
				I3D~\cite{carreira2017quo}$^\ast$  & - & - & 81.76 & 95.16  
				& - & - & 81.97 & 93.79
				& - & - & 57.69 & 81.05\\ 
				TCK~\cite{li2020transferring}  & 83.04 & 93.46 & 83.91 & 93.52  
				& 80.31 & 91.82 & 81.14 & 92.24
				& - & - & - & - \\ 
				BSL~\cite{albanie2020bsl}  & - & - & - & -  
				& - & - & - & -
				& 64.71 & 85.59 & 61.55 & 84.43 \\
				HMA~\cite{hu2021hand} & 73.45 & 89.70 & 74.59  & 89.70 &66.30  & 84.03 & 67.47 & 84.03 & 49.16 & 69.75  & 46.27 & 68.60 \\
                 NLA-SLR-VK32~\cite{zuo2023natural}$^\ddagger$ & - & - &- & - & - & - & - & - &  70.25 &  88.63 &  67.97 &  86.32 \\ 
				SignBERT~(+R)~\cite{hu2021signbert}  & 89.56 & 97.36 & 89.96 & \textbf{97.51}
				& 86.98 & 96.39 & 87.62 & 95.43
				& 71.24 & \textbf{89.12} & 67.96 & 88.40 \\
				BEST~(+R)~\cite{zhao2023best} & 89.56 & 96.96 & \textbf{90.08} & 97.07 & 86.83 & 95.66 & 87.45 & 95.72 & 71.21 & 88.85 & 68.24 & 87.98 \\
				\textit{Ours}~(+R) & \textbf{89.70} & \textbf{97.36} & \underline{89.96} & \underline{97.26} & \textbf{87.27} & \textbf{96.25} & \textbf{88.05} & \textbf{96.24} & \textbf{73.20} & \underline{89.05} & \textbf{70.48} & \textbf{88.54}\\
				\bottomrule
			\end{tabular}
		}
	\end{center}
	\label{msasl}
\end{table*}

\subsection{Datasets and Metrics}
\noindent \textbf{Datasets.} We conduct our experiments on four publicly available benchmarks, \textit{i.e.,} WLASL~\cite{li2020word}, MSASL~\cite{joze2018ms}, NMFs\_CSL~\cite{hu2021global} and SLR500~\cite{huang2018attention}. The entire training sets of all datasets are utilized during the pre-training stage.

WLASL~\cite{li2020word} and MSASL~\cite{joze2018ms} are widely utilized American sign language~(ASL) datasets. For WLASL, it contains 2000 words performed by over 100 signers, which totally consists of 20,183 samples for training and testing. Particularly, it selects the top-$K$ most frequent words with $K=\{100, 300\}$ and organizes them as its two subsets, named WLASL100 and WLASL300. Similarly, MSASL includes a total of 25,513 samples and a vocabulary size of 1000. It also provides two subsets, named MSASL100 and MSASL200, respectively. Importantly, both datasets collect data from unconstrained real-life scenarios and bring more challenges.  

NMFs\_CSL~\cite{hu2021global} and SLR500~\cite{huang2018attention} are both Chinese sign language~(CSL) datasets. Among them, NMFs\_CSL is a large scale dataset with a vocabulary size of 1067 and 32,010 samples, of which 25,608 and 6,402 samples are split into training and testing, respectively. SLR500 is the largest CSL dataset, which consists of 500 words performed by 50 signers. Concretely, it contains a total of 125,000 samples, of which 90,000 and 35,000 samples are utilized for training and testing, respectively. Different from WLASL and MSASL, both datasets are collected from the lab scene. 

\noindent \textbf{Metrics.} For evaluation, we report the classification accuracy, \textit{i.e.,} Top-1~(\textbf{T-1}) and Top-5~(\textbf{T-5}) for the SLR task. We also adopt the Per-class~(\textbf{P-I}) and Per-instance~(\textbf{P-C}) accuracy metrics following~\cite{hu2021global, zhao2023best}, in which \textbf{P-I} denotes the accuracy of the whole test data, while \textbf{P-C} denotes the average accuracy of the sign categories present in the test data.

\subsection{Comparison with State-of-the-art Methods}
In this section, we will compare our proposed method with previous state-of-the-art methods on four public benchmarks. Specifically, we group the previous methods into two categories, \textit{i.e.,} RGB-based and pose-based methods.

\begin{table*}[!t]
	\footnotesize
	\tabcolsep=13pt
	\caption{Comparison with state-of-the-art methods on NMFs-CSL dataset. ``$\dagger$" indicates the model with self-supervised learning, and ``$\ast$" indicates the method utilized for late fusion. }
	\begin{center}
		\resizebox{\textwidth}{!}{
			\begin{tabular}{l|ccc|ccc|ccc}
				\toprule
				\multirow{2}{*}[-0.13in]{Method}      &  \multicolumn{3}{c}{Total} &  \multicolumn{3}{c}{Confusing} &  \multicolumn{3}{c}{Normal}  \\
				\cmidrule(lr){2-4} \cmidrule(lr){5-7} \cmidrule(lr){8-10}
				& T-1 & T-2  & T-5 & T-1 & T-2 &T-5 & T-1 & T-2 & T-5\\ \toprule
				\textbf{Pose-based}  & & & & & & & & & \\
				ST-GCN~\cite{yan2018spatial} & 59.9 & 74.7 & 86.8 & 42.2 & 62.3 & 79.4 & 83.4 & 91.3 & 96.7 \\
				SignBERT~\cite{hu2021signbert}$^\dagger$ & 67.0 & 86.8 & 95.3 & 46.4 & 78.2 & 92.1 & 94.5 & 98.1 & 99.6 \\
				BEST~\cite{zhao2023best}$^\dagger$ & 68.5 & - & 94.4 & 49.0 &   - & 90.3 & 94.6 & -  &  99.7 \\ 
				\textit{Ours} & \textbf{71.7}  & \textbf{89.0} & \textbf{97.0} & \textbf{53.5}& \textbf{81.8}  &  \textbf{92.6} & \textbf{95.9} & \textbf{98.6} &  \textbf{99.9} \\
				\midrule
				\textbf{RGB-based}  & & & & & & & & & \\
				3D-R50~\cite{qiu2017learning}$^\ast$  & 62.1 & 73.2 & 82.9 & 43.1 & 57.9 & 72.4 & 87.4 & 93.4 & 97.0  \\
				I3D~\cite{carreira2017quo}     & 64.4 & 77.9 & 88.0 & 47.3 & 65.7 & 81.8 & 87.1 & 94.3 & 97.3 \\
				TSM~\cite{lin2019tsm}   & 64.5 & 79.5 & 88.7 & 42.9 & 66.0 & 81.0 & 93.3 & 97.5 & 99.0 \\
				GLE-Net~\cite{hu2021global}  & 69.0 & 79.9  & 88.1 & 50.6 & 66.7 & 79.6 & 93.6 & 97.6 & 99.3 \\
				HMA~\cite{hu2021hand} & 64.7 & 81.8 & 91.0 & 42.3 & 69.4 & 84.8 & 94.6 & 98.4 & 99.3 \\
				SignBERT~(+R)~\cite{hu2021signbert}  & 78.4 & 92.0 & 97.3 & 64.3 & 86.5 & 95.4 & 97.4 & 99.3 & 99.9 \\
				BEST~(+R)~\cite{zhao2023best} & 79.2 & -  & 97.1  &  65.5 & - & 95.0 & 97.5  &- & 99.9 \\
				\textit{Ours}~(+R) & \textbf{80.1} & \textbf{93.2} & \textbf{98.3} & \textbf{67.0} & \textbf{89.1} & \textbf{97.0} & \textbf{97.5} & \textbf{99.7} & \textbf{99.9} \\
				\bottomrule
			\end{tabular}
		}
	\end{center}
	\label{NMFs-CSL}
\end{table*}

\begin{table}[t!]
	\footnotesize
	\tabcolsep=32.0pt
	\caption{Comparison on SLR500 dataset.}
	\resizebox{\linewidth}{!}{
		\begin{tabular}{lc}
			\toprule
			Method  &  Accuracy   \\  \toprule
			\textbf{Pose-based} & \\
			ST-GCN~\cite{yan2018spatial} &  90.0 \\ 
			SignBERT~\cite{hu2021signbert}$^\dagger$     & 94.5     \\ 
			BEST~\cite{zhao2023best}$^\dagger$  &  95.4   \\ 
			\textit{Ours} & \textbf{96.3} \\
			\midrule
			\textbf{RGB-based}  & \\
			STIP~\cite{laptev2005space}   &  61.8 \\
			GMM-HMM~\cite{tang2015real} &  56.3 \\
			3D-R50~\cite{qiu2017learning}$^\ast$ &  95.1 \\
			HMA~\cite{hu2021hand} & 95.9 \\
			GLE-Net~\cite{hu2021global}   & 96.8      \\
			SignBERT~(+R)~\cite{hu2021signbert} & 97.6 \\
			BEST~(+R)~\cite{zhao2023best} & 97.7 \\
			\textit{Ours}~(+R) & \textbf{97.7} \\ \bottomrule
		\end{tabular}
	}
	\label{slr500}
\end{table}
\noindent \textbf{WLASL.} As depicted in Tab.~\ref{wlasl}, previous supervised methods, \textit{i.e.,} ST-GCN~\cite{yan2018spatial} and PSLR~\cite{tunga2021pose} show inferior performance compared with RGB-based methods. Although SignBERT~\cite{hu2021signbert} and BEST~\cite{zhao2023best} relieve this issue by employing different self-supervised learning strategies, they still neglect some effective information during temporal modeling. Compared with them, our proposed method improves performance attributed to explicitly exploring motion information and inducing global semantic meanings. Specifically, MASA outperforms BEST~\cite{zhao2023best} with 5.81\%, 5.99\% and 2.81\% Top-1 per-instance accuracy improvement on WLASL100, WLASL300 and WLASL2000, respectively. Notably, our method fused with another simple RGB method still surpasses SignBERT~(+R) and BEST~(+R) with a substantial margin.

\noindent \textbf{MSASL.} As shown in Tab.~\ref{msasl}, the self-supervised methods SignBERT~\cite{hu2021signbert} and BEST~\cite{zhao2023best} have validate the feasibility of masked modeling strategies. Compared with them, our method further mines globally discriminative information to enhance sign language representation learning, and achieve better performance with 2.24\%, 2.65\% and 4.65\% Top-1 per-instance accuracy improvement on MSASL100, MSASL200 and MSASL1000, respectively.

\noindent \textbf{NMFs\_CSL.} As shown in Tab.~\ref{NMFs-CSL}, NMFs\_CSL dataset contains three different settings, \textit{i.e.,} confusing, normal and total. HMA~\cite{hu2021hand} utilizes a hand statistic model to refine the pose and improve the performance. GLE-Net~\cite{hu2021global} is the state-of-the-art method with discriminative clues from global and local views under the supervised learning paradigm. Compared with them and previous self-supervised methods~\cite{hu2021signbert,zhao2023best}, our method still achieves better performance with 71.7\% top-1 accuracy.

\begin{table}[t]
	\centering
	\tabcolsep=6pt
	\caption{Effect of different masked strategies and objectives on WLASL dataset.}
	\begin{center}
		\resizebox{\linewidth}{!}{
			\begin{tabular}{cccccccc}
				\toprule
				\multirow{2}{*}[-0.05in]{\makecell[c]{Mask.}} & \multirow{2}{*}[-0.05in]{\makecell[c]{Obj.}} & \multicolumn{2}{c}{WLASL100} & \multicolumn{2}{c}{WLASL300} & \multicolumn{2}{c}{WLASL2000} \\
				\cmidrule(lr){3-4} \cmidrule(lr){5-6} \cmidrule(lr){7-8}
				& & P-I & P-C &  P-I &  P-C &  P-I &  P-C\\ \midrule
				Random & Joint  & 77.13  & 77.52  & 67.12  & 67.84 & 43.62 & 40.52   \\
				Motion & Joint  & 80.62  & 80.63  & 70.32  & 70.72  & 47.22 & 44.30  \\
				Random & Motion  & 81.40 & 81.72  &  71.86 &  72.41 & 47.78 & 45.40  \\
				Motion & Motion & \textbf{83.72} & \textbf{84.47} & \textbf{73.65} &\textbf{74.33}  & \textbf{49.06} &\textbf{46.91} \\ \bottomrule          
			\end{tabular}
		}
	\end{center}
	\label{different_mask_objective}
\end{table}
\begin{table}[t]
	\centering
	\tabcolsep=7pt
	\caption{Effect of interval length $k$ in MA on WLASL dataset.}
	\begin{center}
		\resizebox{\linewidth}{!}{
			\begin{tabular}{ccccccc}
				\toprule
				\multirow{2}{*}[-0.05in]{\makecell[c]{ Interval $k$}} & \multicolumn{2}{c}{WLASL100} & \multicolumn{2}{c}{WLASL300} & \multicolumn{2}{c}{WLASL2000} \\
				\cmidrule(lr){2-3} \cmidrule(lr){4-5} \cmidrule(lr){6-7}
				& P-I & P-C &  P-I &  P-C &  P-I &  P-C\\ \midrule
				1  & 81.40 &  81.80 & 69.16 & 69.64 &  47.64 &  45.17  \\
				3  &  \textbf{83.72} & \textbf{84.47} & \textbf{73.65} &\textbf{74.33}  & \textbf{49.06} &\textbf{46.91}  \\
				5 &  81.01 &  81.55 & 72.31 & 72.51 & 47.71  & 45.36  \\ 
				7  &  79.46 &  79.72 &  70.81 & 70.99 & 45.03 & 42.85 \\	\bottomrule        
			\end{tabular}
		}
	\end{center}
	\label{motion interval}
\end{table}

\begin{table}[t]
	\centering
	\tabcolsep=7pt
	\caption{Effect of different components in our proposed framework on WLASL dataset.}
	\begin{center}
		\resizebox{0.5\textwidth}{!}{
			\begin{tabular}{cccccccc}
				\toprule
				\multirow{2}{*}[-0.05in]{\makecell[c]{MA}} & \multirow{2}{*}[-0.05in]{\makecell[c]{SA}} & \multicolumn{2}{c}{WLASL100} & \multicolumn{2}{c}{WLASL300} & \multicolumn{2}{c}{WLASL2000} \\
				\cmidrule(lr){3-4} \cmidrule(lr){5-6} \cmidrule(lr){7-8}
				& & P-I & P-C &  P-I &  P-C &  P-I &  P-C\\ \midrule
				\checkmark &  & 81.01 & 81.05  & 69.76 & 70.31 & 47.64 & 45.53   \\
				& \checkmark & 74.42 & 75.80 & 66.47 & 67.37 & 43.61 & 41.34 \\
				\checkmark & \checkmark & \textbf{83.72} & \textbf{84.47} & \textbf{73.65} &\textbf{74.33}  & \textbf{49.06} &\textbf{46.91} \\ \bottomrule          
			\end{tabular}
		}
	\end{center}
	\label{different components}
\end{table}

\noindent \textbf{SLR500.} As shown in Tab.~\ref{slr500}, compared with deep learning methods, STIP~\cite{laptev2005space} and GMM-HMM~\cite{tang2015real} with hand-craft features show poor performance due to the inferior representations. Among deep learning methods, self-supervised methods~\cite{hu2021signbert,zhao2023best} learn the powerful capacity of representation learning compared to conventional supervised methods~\cite{yan2018spatial,hu2021global,hu2021hand}. Compared with them, our method still achieves new state-of-the-art performance.

\subsection{Ablation Study}
In this section, we conduct several crucial ablation experiments to validate the effectiveness of our method, For fair comparison, we conduct experiments on the WLASL~\cite{li2020word} dataset and its two subsets, named WLASL2000, WLASL300 and WLASL100, respectively.

\noindent \textbf{Different Components.} We validate the effectiveness of different components in our proposed method, as shown in Tab.~\ref{different components}.  The MA utilizes a masked modeling strategy to capture motion information among adjacent frames, while the SA aims to learn instance discriminative representation of the intact sign pose sequence via aligning features. Both components endow our model with different temporal perceptions. It is observed that the integrity of both components achieves better performance than adopting any one of them individually, which proves the feasibility of exploring more temporal information for sign language learning.

\noindent \textbf{Different Masked Modeling Strategies.} As shown in Tab.~\ref{different_mask_objective}, the first row denotes the baseline method with random masked strategy and the prevalent objective with predicting static  joints of masked frames.  It is notable that compared with this baseline, our proposed motion-aware masked strategy and objective bring remarkable performance gain with over 3\% accuracy improvement  on all subsets of WLASL dataset.
Moreover, the proposed objective shows more effectiveness than the proposed masked strategy, which inclines the importance of explicitly mining dynamic motion cues.
When both designs are utilized, it reaches the best performance, with 6.59\%, 6.53\% and 5.44\% Top-1 per-instance accuracy improvement on WLASL100, WLASL300 and WLASL2000, respectively.
\begin{figure}[t]
	\centering
	\includegraphics[width=\linewidth]{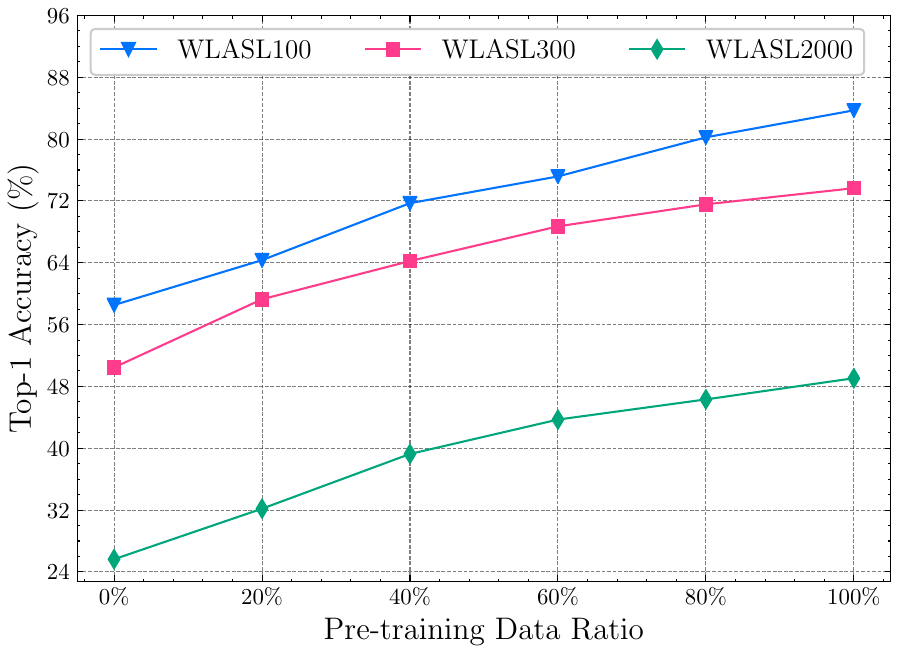} 
	\caption{Effect of the data scale during pre-training on WLASL dataset. The horizontal coordinate denotes the data ratio, while the vertical coordinate denotes recognition accuracy.}
	\label{data ratio}
\end{figure}

\begin{table}[t]
	\centering
	\tabcolsep=8pt
	\caption{Effect of different mask ratio in our proposed framework on WLASL dataset.}
	\renewcommand\arraystretch{1}
	\begin{center}
		\resizebox{0.5\textwidth}{!}{
			\begin{tabular}{cccccccc}
				\toprule
				\multirow{2}{*}[-0.05in]{\makecell[c]{Ratio}} & \multicolumn{2}{c}{WLASL100} & \multicolumn{2}{c}{WLASL300} & \multicolumn{2}{c}{WLASL2000} \\
				\cmidrule(lr){2-3} \cmidrule(lr){4-5} \cmidrule(lr){6-7}
				& P-I & P-C &  P-I &  P-C &  P-I &  P-C \\ \midrule
				0\% & 58.53 & 58.78  & 50.45 & 50.87 & 25.64 & 23.21 \\
				10\% & 65.50 & 65.65 & 59.28 & 59.93 & 31.13 & 28.82 \\
				30\% & 71.32 & 71.47 & 65.12 & 65.98 & 39.19 & 38.28 \\
				50\% & 74.81 & 75.43 & 67.66 & 68.37 & 42.84 & 40.49 \\
				70\% & 79.84 & 80.30 & 71.11 & 71.83 & 46.46 & 44.42 \\
				90\% & \textbf{83.72} & \textbf{84.47} & \textbf{73.65} &\textbf{74.33}  & \textbf{49.06}  &\textbf{46.91} \\ \bottomrule      
			\end{tabular}
		}
	\end{center}
	\label{mask_ratio}
\end{table}
\noindent \textbf{Interval Length $k$.} In Tab.~\ref{motion interval}, we further explore the impact of different interval length $k$ for computing motion residuals between two frames. It is observed that a smaller value leads to the simplification of the model learning, while a larger value makes the model difficult to converge, and ultimately both lead to sub-optimal solutions. Finally, we set $k$ to 3 as the default setting due to its best performance on WLASL dataset.

\noindent \textbf{Pre-training Data Scale.} As shown in Fig.~\ref{data ratio}, we investigate the effect of pre-training data scale. The 0\% denotes that our framework directly performs the SLR task without pre-training. It is observed that the performance gradually increases with the increment of pre-training data scale. Importantly, we evenly select the same proportion of data from each dataset. The result validates that our method is suitable for pre-training with large-scale data.

\begin{table}[t]
	\centering
	\tabcolsep=8pt
	\caption{Effect of objective weight in our proposed framework on WLASL dataset.}
	\renewcommand\arraystretch{1}
	\begin{center}
		\resizebox{0.5\textwidth}{!}{
			\begin{tabular}{cccccccc}
				\toprule
				\multirow{2}{*}[-0.05in]{\makecell[c]{$\lambda_s$}} & \multicolumn{2}{c}{WLASL100} & \multicolumn{2}{c}{WLASL300} & \multicolumn{2}{c}{WLASL2000} \\
				\cmidrule(lr){2-3} \cmidrule(lr){4-5} \cmidrule(lr){6-7}
				& P-I & P-C &  P-I &  P-C &  P-I &  P-C \\ \midrule
				0.025 & 79.84 & 80.13 & 71.26 & 71.74 & 46.28 & 44.21 \\
				0.05 & \textbf{83.72} & \textbf{84.47} & \textbf{73.65} &\textbf{74.33}  & \textbf{49.06}  &\textbf{46.91} \\
				0.075 & 81.40 & 82.10 & 70.81 & 71.53 & 47.67 & 45.23 \\
				0.1 & 79.84 & 80.30 & 69.91 & 70.58 & 46.46 & 44.42 \\ \bottomrule          
			\end{tabular}
		}
	\end{center}
	\label{loss_weight}
\end{table}

\begin{table}[t]
	\centering
	\tabcolsep=8pt
	\caption{Effect of the hyper-parameter $\epsilon_m$ in our proposed framework on WLASL dataset.}
	\renewcommand\arraystretch{1}
	\begin{center}
		\resizebox{0.5\textwidth}{!}{
			\begin{tabular}{cccccccc}
				\toprule
				\multirow{2}{*}[-0.05in]{\makecell[c]{$\epsilon_m$}} & \multicolumn{2}{c}{WLASL100} & \multicolumn{2}{c}{WLASL300} & \multicolumn{2}{c}{WLASL2000} \\
				\cmidrule(lr){2-3} \cmidrule(lr){4-5} \cmidrule(lr){6-7}
				& P-I & P-C &  P-I &  P-C &  P-I &  P-C \\ \midrule
				0 & 79.57 & 80.23  & 70.15 & 71.26 & 46.21 & 43.51 \\
				5 & \textbf{83.72}  & \textbf{84.47} & \textbf{73.65} & \textbf{74.33} & \textbf{49.06} & \textbf{46.91} \\
				10 & 82.17 & 83.09 & 72.69 & 73.34 & 48.52 & 46.23 \\
				15 & 80.78 & 81.45 & 71.02 & 72.28 & 46.36 & 44.38  \\ \bottomrule      
			\end{tabular}
		}
	\end{center}
	\label{hyper_epsilon_m}
\end{table}

\begin{table}[t]
	\centering
	\tabcolsep=8pt
	\caption{Effect of the hyper-parameter $\delta$ in our proposed framework on WLASL dataset.}
	\renewcommand\arraystretch{1}
	\begin{center}
		\resizebox{0.5\textwidth}{!}{
			\begin{tabular}{cccccccc}
				\toprule
				\multirow{2}{*}[-0.05in]{\makecell[c]{$\delta$}} & \multicolumn{2}{c}{WLASL100} & \multicolumn{2}{c}{WLASL300} & \multicolumn{2}{c}{WLASL2000} \\
				\cmidrule(lr){2-3} \cmidrule(lr){4-5} \cmidrule(lr){6-7}
				& P-I & P-C &  P-I &  P-C &  P-I &  P-C \\ \midrule
				0.1 & 80.13 & 80.79  & 69.92 & 70.43 & 47.12 & 44.69 \\
				0.5 & \textbf{83.72}  & \textbf{84.47} & \textbf{73.65} & \textbf{74.33} & \textbf{49.06} & \textbf{46.91} \\
				0.9 & 79.64 & 80.37 & 69.35 & 70.27 & 48.52 & 46.23 \\ \bottomrule      
			\end{tabular}
		}
	\end{center}
	\label{hyper_delta}
\end{table}

\noindent \textbf{Mask Ratio.} We conduct the experiment to investigate the impact of different mask ratio in MA as shown in Tab.~\ref{mask_ratio}. The first row denotes that our method directly performs the SLR task without pre-training as a baseline. It is observed that the performance gradually improves as the mask ratio increases. Our framework could adapt to higher mask ratio with our designed mased strategy and effectively models dynamic relations in local regions. Compared with the baseline, our method achieves 83.72\%, 73.65\% and 49.06\% Top-1 per-instance accuracy on WLASL dataset.

\noindent \textbf{Objective Weight.} We further research the impact of the objective weight in our proposed method. Considering that semantic consistency loss $\mathcal{L}_s$ as an auxiliary loss aims to endow our model with instance awareness, we select the lower value for this objective. As shown in Tab.~\ref{loss_weight}, we find the proper value of hyper-parameter $\lambda_s$ as 0.05 based on the performance on WLASL dataset.

\noindent \textbf{Hyper-parameter Sensitivity Analysis.} We provide more sensitivity analysis on hyper-parameters in our proposed framework. 1) $\epsilon_c$ is a threshold to truncate low confidence. We observe the estimated poses from MMPose with different confidences from 0.1 to 1.0 and find that the estimated error is beyond the acceptable range when the confidence level is below 0.4, so we set $\epsilon_c$ to 0.4. 2) $\epsilon_m$ is utilized to compute the ratio of keypoints with valid motion. We conduct the experiment on $\epsilon_m$ in the Tab.~\ref{hyper_epsilon_m}. It is observed that the performance is optimized when $\epsilon_m$ is set to 5. 3) We also conduct the experiment on the threshold $\delta$, as shown in the Tab.~\ref{hyper_delta}. The lower threshold introduces more inaccurate motion noise, while the higher threshold truncates some valid motion information. Weighing these reasons, we set the threshold $\delta$ as 0.5.

\begin{table}[t]
	\centering
	\tabcolsep=8pt
	\caption{Effect of different qualities of input data in our proposed framework on WLASL dataset. $\sigma$ denotes the intensity of the Gaussian noise that controls data quality.}
	\renewcommand\arraystretch{1}
	\begin{center}
		\resizebox{0.5\textwidth}{!}{
			\begin{tabular}{cccccccc}
				\toprule
				\multirow{2}{*}[-0.05in]{\makecell[c]{$\sigma$}} & \multicolumn{2}{c}{WLASL100} & \multicolumn{2}{c}{WLASL300} & \multicolumn{2}{c}{WLASL2000} \\
				\cmidrule(lr){2-3} \cmidrule(lr){4-5} \cmidrule(lr){6-7}
				& P-I & P-C &  P-I &  P-C &  P-I &  P-C \\ \midrule
				0 & \textbf{83.72}  & \textbf{84.47} & \textbf{73.65} & \textbf{74.33} & \textbf{49.06} & \textbf{46.91} \\
				5 & 83.55 & 84.21 & 73.47 & 74.18 & 48.95 & 46.74 \\
				10 & 83.14 & 83.87 & 73.19 & 73.92 & 48.61 & 46.42 \\ 
				15 & 80.24 & 81.19 & 70.23 & 70.47 & 45.26 & 43.29 \\ 
				20 & 71.93 & 73.26 & 62.12 & 63.08 & 42.15 & 40.35 \\ \bottomrule      
			\end{tabular}
		}
	\end{center}
	\label{hyper_sigma}
\end{table}

\noindent \textbf{Different Qualities of Input Data.} Since our method utilized the pose data extracted from an off-the-shelf human pose estimator, we analyze the sensitivity on the quality of input data. We simulate different qualities of input data by adding Gaussian noise with different intensities $\sigma$ to the input skeleton points. The noise is formulated as $n = N(0, 1) * \sigma$, where $N(0, 1)$ denotes the standard normal distribution. The coordinates of the pose data range from -128 to 128. For example, if $\sigma = 10$, the average offset caused by noise is roughly $\pm 20$. We utilize pose data with different qualities as the input and report the corresponding result in the Tab.~\ref{hyper_sigma}. It is observed that as the noise gradually increases, the accuracy gradually decreases. Moreover, if the noise is maintained in a small range, the accuracy does not degrade significantly.

\subsection{Limitation}
Our proposed method relies on the detection accuracy of the off-the-shelf pose estimator.
Since the quality of the extracted keypoints of the pose estimator is somewhat sensitive to the video resolution and frame rate, our proposed method possibly shows limited improvement in low-resolution or human-motion blurred scenes.
In particular, the occurrence of mutual occlusion during hand interaction also poses a significant challenge to learn sign language representations.
Although we introduce the pose confidence into our framework to alleviate this dilemma, the representation learning may be potential for improvement with more accurate sign pose data.

In addition, our proposed motion-aware masked autoencoder predicts the movement of the human keypoints at a fixed interval length. However, signers often exhibit different frequencies in sign language videos depending on their personal habitation. Therefore, the motion with a fixed interval length maybe struggle with complex scenarios where motion occurs at multiple scales. In contrast, the multi-scale motion-aware strategy could introduce a more delicate understanding of motion dynamics by incorporating information from multiple temporal scales. Incorporating multi-scale motion-aware strategy is potential for capturing more dynamic cues among sign pose sequences.

\section{Conclusion}
In this paper, we propose a novel self-supervised pre-training framework, named MASA, to learn more effective sign language representation. MASA consists of two primary components, including a motion-aware masked autoencoder~(MA) and a momentum semantic alignment module~(SA). In MA, we design a motion-aware masked strategy and an objective function with motion prediction to explicitly mine motion information in sign pose data. Moreover, considering the significance of global semantic meaning in lexical signs, we employ SA to constrain the semantic consistency of positive views by aligning their embeddings in a shared embedding space. Consequently, we fully leverage rich motion cues and global instance discriminativeness to enhance the representative capability of our framework. Extensive experiments are conducted to validate the effectiveness of our proposed framework, resulting in new state-of-the-art performance on four benchmarks.

Despite achieving promising results, there are still some directions worth exploring. 1) Currently, we employ a general-purpose human pose estimator to conveniently acquire sign pose data. However, such estimators typically lack precision in capturing the detailed interacting hand gestures involved in the sign language domain, thereby imposing limitations on the model's representation capacity. Thus, investigating approaches for acquiring high-quality sign language pose data emerges as a meaningful research direction. 2) Compared to explicitly modeling motion information from sparse pose sequences, effectively capturing dynamic cues from informative RGB sign language videos is more efficient and effective in real-world scenarios. Hence,  it is also desirable to extend motion-aware self-supervised pre-training to RGB data. 
3) Due to the limitations of fixed frame interval length for modeling various motion information, it is also valuable to explore the multi-scale motion-aware strategy with semantic alignment in the sign language domain.

\bibliographystyle{ieee_fullname}
\bibliography{main}

\newpage

\begin{IEEEbiography}[{\includegraphics[width=1in,height=1.25in,clip,keepaspectratio]{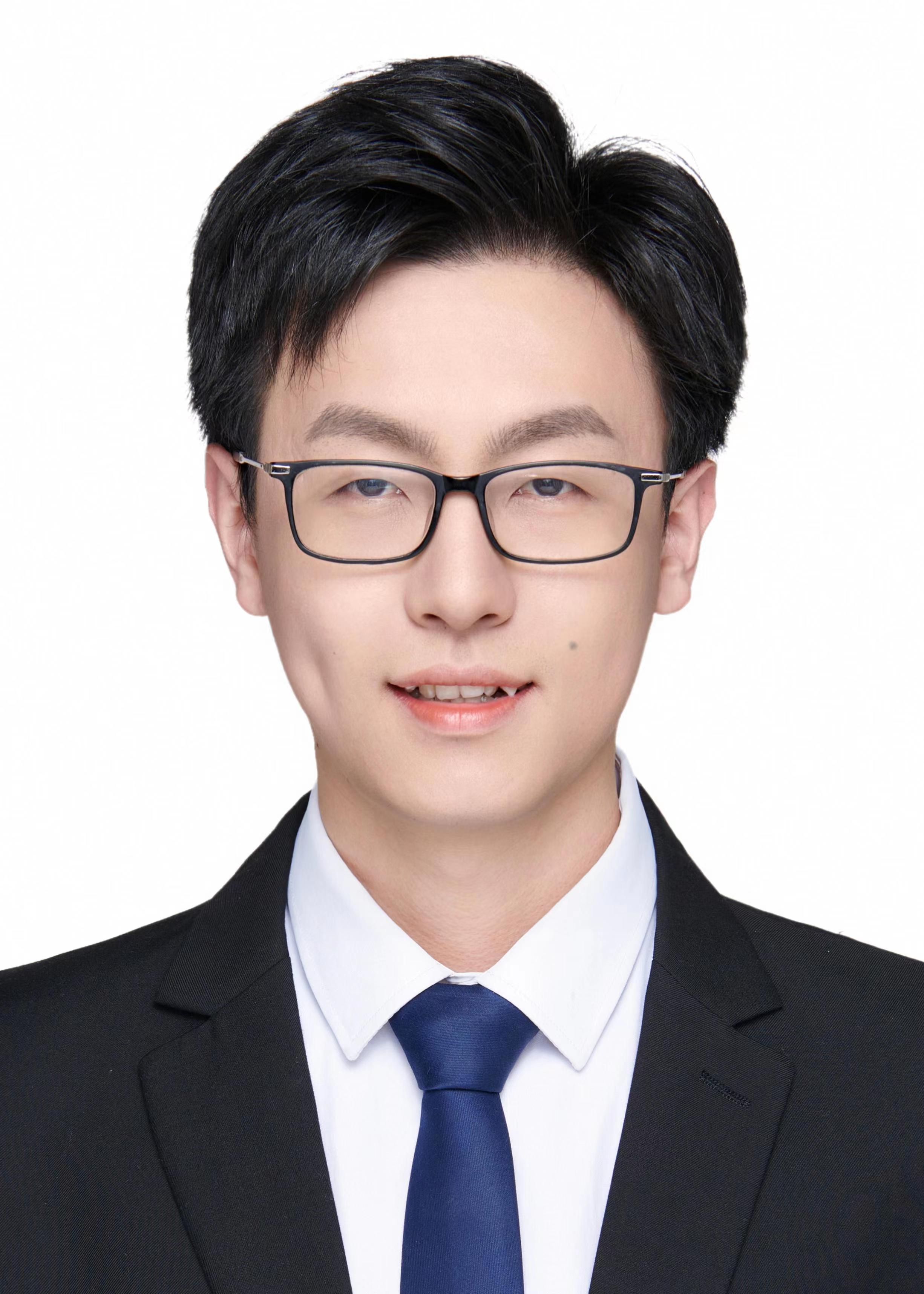}}]{Weichao Zhao} received B.E. degree in electronic information engineering from the University of Science and Technology of China~(USTC), and is currently pursuing the Ph.D. degree in data science with the School of Data Science from USTC.
His research interests include sign language understanding, self-supervised pre-training, multimodal representation learning and computer vision.
\end{IEEEbiography}

\begin{IEEEbiography}[{\includegraphics[width=1in,height=1.25in,clip,keepaspectratio]{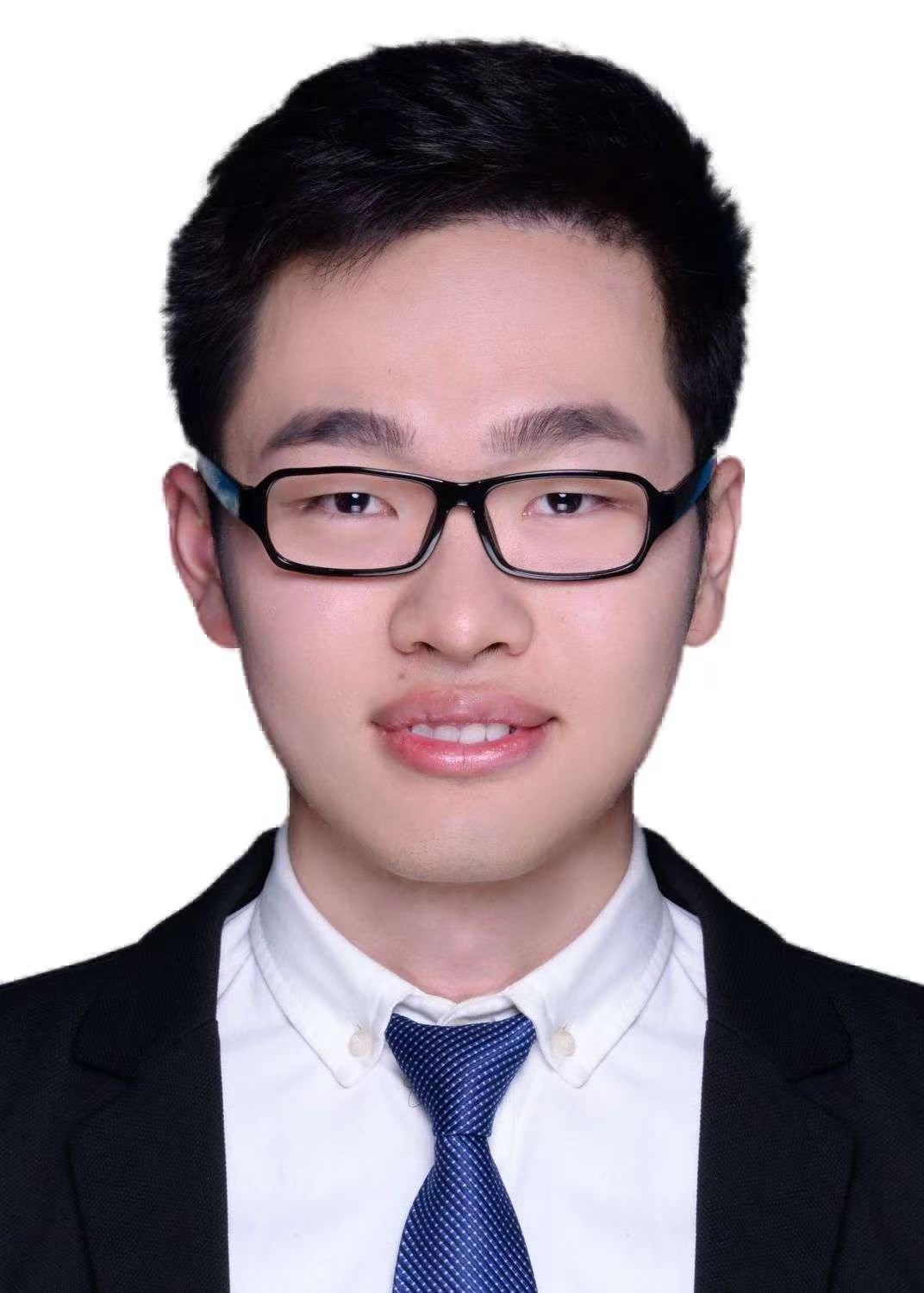}}]{Hezhen Hu} is a postdoctoral researcher at The University of Texas at Austin. He received the Ph.D. degree in information and communication engineering with the Department of Electronic Engineering and Information Science, from the University of Science and Technology of China~(USTC), in 2023. His research interests include sign language understanding, self-supervised pre-training, human-centric visual understanding, and computer vision.
\end{IEEEbiography}

\begin{IEEEbiography}[{\includegraphics[width=1in,height=1.25in,clip,keepaspectratio]{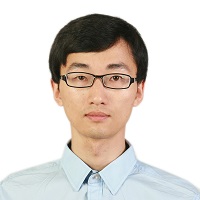}}]{Wengang Zhou} (S'20) received the B.E. degree in electronic information engineering from Wuhan University, China, in 2006, and the Ph.D. degree in electronic engineering and information science from the University of Science and Technology of China (USTC), China, in 2011. From September 2011 to September 2013, he worked as a postdoc researcher in Computer Science Department at the University of Texas at San Antonio. He is currently a Professor at the EEIS Department, USTC. 

His research interests include multimedia information retrieval, computer vision, and computer game. In those fields, he has published over 100 papers in IEEE/ACM Transactions and CCF Tier-A International Conferences. He is the winner of National Science Funds of China (NSFC) for Excellent Young Scientists. He is the recepient of the Best Paper Award for ICIMCS 2012. He received the award for the Excellent Ph.D Supervisor of Chinese Society of Image and Graphics (CSIG) in 2021, and the award for the Excellent Ph.D Supervisor of Chinese Academy of Sciences (CAS) in 2022. He won the First Class Wu-Wenjun Award for Progress in Artificial Intelligence Technology in 2021. He served as the publication chair of IEEE ICME 2021 and won 2021 ICME Outstanding Service Award. He is currently an Associate Editor and a Lead Guest Editor of IEEE Transactions on Multimedia. 
\end{IEEEbiography}

\begin{IEEEbiography}[{\includegraphics[width=1in,height=1.25in,clip,keepaspectratio]{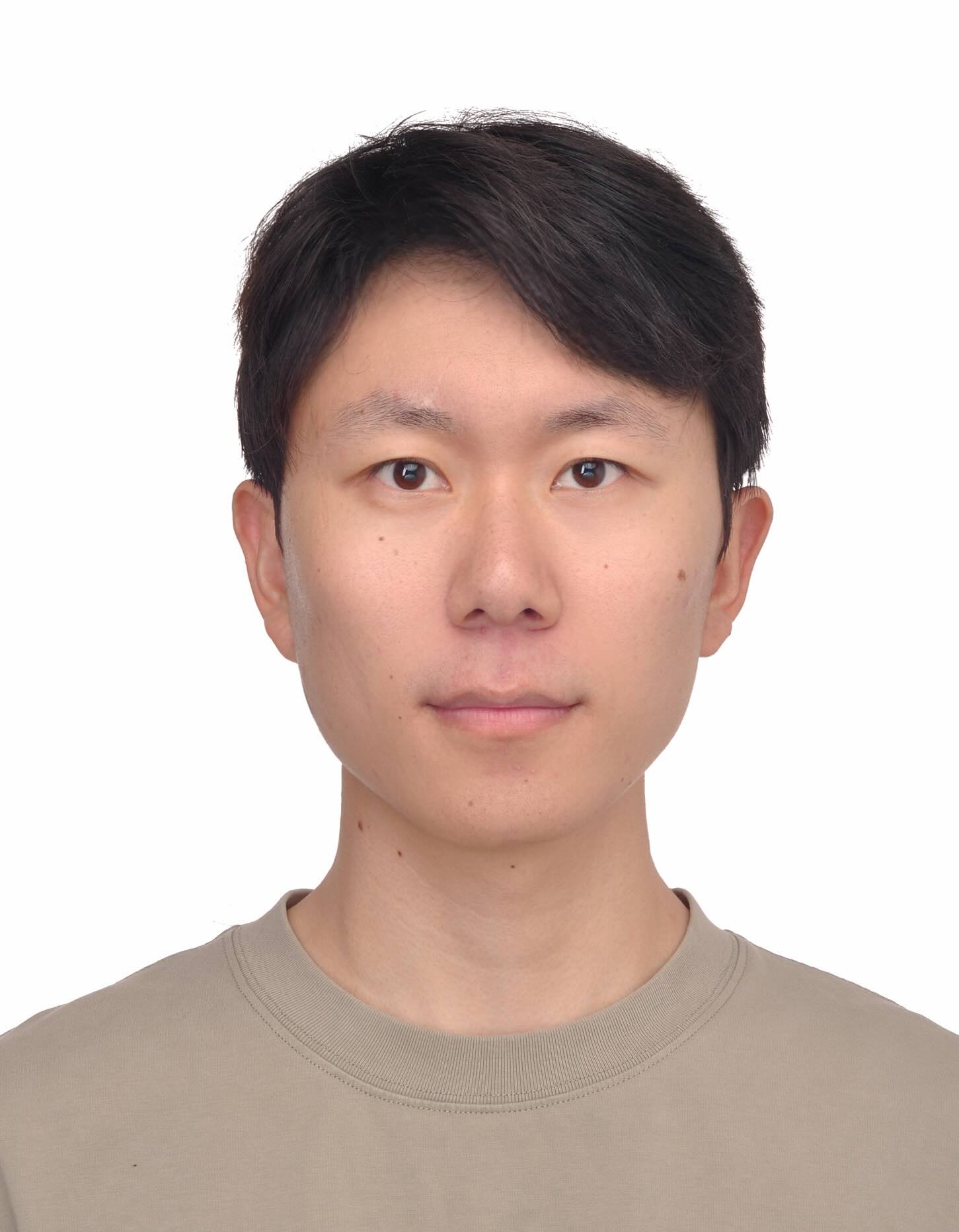}}]{Yunyao Mao} received the B.E. degree in information security from the University of Science and Technology of China (USTC), China, in 2020. He is currently pursuing the Ph.D. degree with the Department of Electronic Engineer and Information Science, University of Science and Technology of China (USTC). His research interest is computer vision. His current research work is focused on human activity understanding.
\end{IEEEbiography}

\begin{IEEEbiography}[{\includegraphics[width=1in,height=1.25in,clip,keepaspectratio]{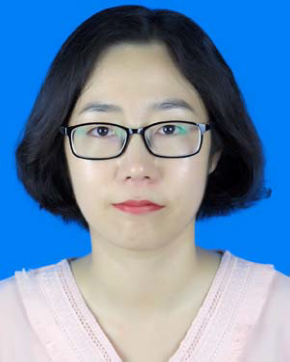}}]{Min Wang} received the BE, and PhD degrees in electronic information engineering from the University of Science and Technology of China (USTC), in 2014 and 2019, respectively. From July 2019 to 2020, she worked with Huawei Noah’s Ark Lab. She is working with the Institute of Artificial Intelligence, Hefei Comprehensive National Science Center. Her
current research interests include multimedia information retrieval and computer vision.
\end{IEEEbiography}

\begin{IEEEbiography}[{\includegraphics[width=1in,height=1.25in,clip,keepaspectratio]{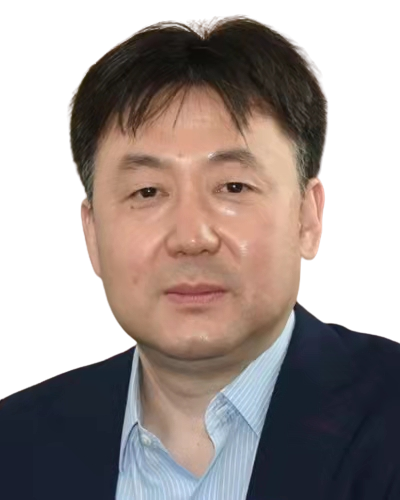}}]{Houqiang Li} (S'12, F'21) received the B.S., M.Eng., and Ph.D. degrees in electronic engineering from the University of Science and Technology of China, Hefei, China, in 1992, 1997, and 2000, respectively, where he is currently a Professor with the Department of Electronic Engineering and Information Science. 
	
His research interests include image/video coding, image/video analysis, computer vision, reinforcement learning, etc.. He has authored and co-authored over 200 papers in journals and conferences. He is the winner of National Science Funds (NSFC) for Distinguished Young Scientists, the Distinguished Professor of Changjiang Scholars Program of China, and the Leading Scientist of Ten Thousand Talent Program of China. He is the associate editor (AE) of IEEE TMM, and served as the AE of IEEE TCSVT from 2010 to 2013. He served as the General Co-Chair of ICME 2021 and the TPC Co-Chair of VCIP 2010. He received the second class award of China National Award for Technological Invention in 2019, the second class award of China National Award for Natural Sciences in 2015, and the first class prize of Science and Technology Award of Anhui Province in 2012. He received the award for the Excellent Ph.D Supervisor of Chinese Academy of Sciences (CAS) for four times from 2013 to 2016. He was the recipient of the Best Paper Award for VCIP 2012, the recipient of the Best Paper Award for ICIMCS 2012, and the recipient of the Best Paper Award for ACM MUM in 2011.
\end{IEEEbiography}

\end{document}